\documentclass{article}

\usepackage{template/spconf}

\usepackage[T1]{fontenc}
\usepackage[utf8]{inputenc}
\usepackage[english]{babel}

\usepackage{newtxtext}

\usepackage{mathtools} 
\usepackage{amsthm}

\usepackage[nonewtxmathopt]{newtxmath} 
\usepackage[cal=cm,bb=ams,frak=pxtx]{mathalpha}

\PassOptionsToPackage{hyphens}{url}
\usepackage[hyphens]{url}
\usepackage[hidelinks]{hyperref}

\usepackage{graphicx}
\usepackage{caption}
\usepackage{tikz}
\usetikzlibrary{hobby,decorations.markings}
\usepackage{pgfplots}
\pgfplotsset{compat=1.17}

\usepackage{booktabs}
\usepackage{microtype}
\usepackage{xcolor}
\usepackage{enumitem}
\setitemize{itemsep=0pt,parsep=0pt,topsep=0pt}
\setenumerate{itemsep=0pt,parsep=0pt,topsep=0pt}
\usepackage[ruled,vlined]{algorithm2e}

\usepackage{xfrac}

\newcommand{\diff}{\mathop{}\!\mathrm{d}}

\DeclareMathOperator{\E}{\mathbb{E}}

\DeclareMathOperator*{\argmin}{arg\,min}

\DeclareMathOperator*{\supp}{supp}

\newcommand{\R}{\mathbb{R}}

\renewcommand{\brace}[1]{\left\{ #1 \right\}}
\newcommand{\bracket}[1]{\left[ #1 \right]}

\newcommand{\paren}[1]{\left( #1 \right)}
\newcommand{\midvert}{\,\middle\vert\,}

\newcommand{\norm}[1]{\left\| #1 \right\|}
\newcommand{\scap}[2]{\left\langle #1, #2 \right\rangle}

\newcommand{\X}{\mathcal{X}}
\newcommand{\Y}{\mathcal{Y}}

\renewcommand{\epsilon}{\varepsilon}
\renewcommand{\phi}{\varphi}

\theoremstyle{plain}

\usepackage[normalem]{ulem}

\title{On minimal variations for unsupervised representation learning}
\name{Vivien Cabannes \qquad Alberto Bietti\qquad Randall Balestriero}
\address{Meta AI}

\begin{document}

\maketitle

\begin{abstract}
Unsupervised representation learning aims at describing raw data efficiently to solve various downstream tasks.
It has been approached with many techniques, such as manifold learning, diffusion maps, or more recently self-supervised learning.
Those techniques are arguably all based on the underlying assumption that target functions, associated with future downstream tasks, have low variations in densely populated regions of the input space.
Unveiling minimal variations as a guiding principle behind unsupervised representation learning paves the way to better practical guidelines for self-supervised learning algorithms.

\end{abstract}


\section{Introduction}

Data is everywhere, but it is often too unstructured or high dimensional to leverage classical statistics on their raw form. 
Recent advances in machine learning have succeeded in exploiting parts of the millions terabytes of unlabeled data contained on the internet.
This was achieved by creating self-supervised tasks to be solved by the machine, inciting it to learn good representations of text \cite{devlin_bert_2019,chowdhery_palm_2022}. 
Those ``foundational'' representations are now being leveraged to solve several ``downstream'' tasks on languages \cite{brown_language_2020}.
Similar developments have been made on other high-dimensional data such as images, videos or audio speeches \cite{chen_simple_2020,caron_unsupervised_2020,radford_robust_2022}.
Despite their rapid progress, the training of self-supervised learning (SSL) models remains challenging and lacks theoretical foundations.

Learning without supervision has been historically referred to as unsupervised learning.
While at first sights, the literature bodies on unsupervised learning and self-supervised learning seem relatively disjoint, connections have been made between the two \cite{haochen_provable_2021,balestriero_contrastive_2022}.
This work provides further insights on their links through the concept of minimal variations, detailed in Section \ref{sec:theory}.
Theory is verified on synthetic experiments in Section~\ref{sec:exp}.
This understanding could be leveraged in future work to improve SSL algorithms in practical settings with limited compute resources.

\section{Setting and context}
In the following, $\X$ shall be a Hilbert space ({\em i.e.} endowed with a scalar product) and $\Y$ an output space.
A distribution $\rho_X$ is assumed to have generated a dataset $(X_i)_{i\in[n]}$ of independent samples $X_i \sim \rho_X$ for $i\in[n]$.\footnote{The set $\brace{1, \cdots, n}$ is denoted $[n]$.}
Our goal is to find a representation $\phi:\X\to\R^p$ for a small $p$, such that for relevant downstream distributions $\rho$ on pairs of input/output and loss functions $\ell:\Y\times\Y\to\R$, one can efficiently minimize the subsequent population risk
\begin{equation}
    {\cal R}(f; p, \ell) = \E_{(X,Y)\sim \rho}\bracket{\ell(f(X), Y)},
\end{equation}
based on i.i.d. samples $(X_i, Y_i)$.
More exactly, the optimal functions $f$ should easily be approached under the form $f = g\circ \phi$ for $g$ in a small class of functions. 
Typically, $g$ would be a linear function, a.k.a. a linear probe.
Reducing the search of $f:\X\to\Y$ in a potentially big function space to the search of $g:\R^p\to\Y$ in a much smaller one, will drastically improve sample efficiency \cite{vapnik_nature_1995}.

To learn the representation $\phi$, SSL leverages augmentations of data.
It defines $t:\X\times\Xi\to\X$ a transformation parameterized by $\xi\in\Xi$.
For example, $\Xi$ could be $\X$ and $t(X, \xi) = X + \xi$.
Assuming that transformations $t(X, \xi)$ do not fundamentally change the semantic of the input, any pairs of augmented and original points should be close in the features space $\phi(\X)$.
This is put in equations through the minimization of the variational quantity
\begin{equation}
    \label{eq:ssl}
    \E_{X,\xi}\bracket{d(\phi(t(X, \xi)), \phi(X))},
\end{equation}
for $d$ a notion of similarity in $\R^p$, {\em e.g.} the square loss $d(x, x') = \norm{x-x'}^2$.
In practice, $\phi$ is often taken as a neural network, and its learning is conducted through the optimization of its parameters.
Equation \eqref{eq:ssl} is trivially minimized by setting $\phi$ to a constant.
To avoid such a ``collapse'' phenomenon, one should encourage diversity in the representation, for instance by using the constraint
\begin{equation}
    \label{eq:const}
    \E_X[\phi(X)\phi(X)^\top] = I.
\end{equation}
Classical self-supervised techniques such as SimCLR \cite{chen_simple_2020}, Barlow Twins \cite{zbontar_barlow_2021} and VICReg \cite{bardes_vicreg_2022} can be understood as implementing different specifications of such a scheme \cite{balestriero_contrastive_2022} (respectively, $d$ would be the cosine similarity, some cross-correlation measure, and the square hinge loss).

\section{Minimal variations}
\label{sec:theory}

\subsection{Classical hypothesis}
This section reviews classical assumptions about the nature of downstream tasks with respect to the input distribution $\rho_X$.
It shows how those assumptions are related to the idea that future target functions have low variations on highly populated regions of the input~space.

Often praised in semi-supervised learning setting, the {\em cluster assumption} states that the support of $\rho_X$ have several connected components and that downstream classification tasks are likely to respect this structure, {\em i.e.} labels shall be constant over each connected component a.k.a.~cluster \cite{rigollet_generalization_2007}.
In other terms, one expects the decision boundary\footnote{For a predictor $f:\X\to\Y$, the input space $\X$ is partitioned into decision regions $\X_y = \brace{x\midvert f(x) = y}$ indexed by $y\in\Y$, ``decision boundaries'' refers to the boundaries of those regions.} between classes to be situated in regions of the input space where there is no density.
Yet, on big or poorly curated datasets, classes might not be separated by no-density regions.
In such a setting, the cluster assumption is relaxed as the {\em low-density separation hypothesis}, assuming that downstream decision boundaries will fall in low-density regions ({\em i.e.} where $\rho_X$ is small). 
For example, in a balanced binary classification problem where $\Y=\brace{-1, 1}$ and $\diff \rho\paren{x\midvert y} \propto \exp(-\norm{x+y\mu}/\sigma^2) \diff x$, the optimal decision boundary is the hyperplane $\mu^\perp$ which does minimize the value of $\rho_X(A)$ for any hyperplane $A$ that cross $[-\mu, +\mu]$.

In regression settings, it is often assumed that the downstream target functions will be smooth on densely populated regions of the input space \cite{engelen_surver_2020}.
Variations of functions are measured through quantities such as
\begin{equation}
    \label{eq:crit}
    {\cal J}(f) = \E\bracket{\norm{\nabla f(X)}^q},
\end{equation}
for $q=1$ (total variation), $q=2$ (Dirichlet energy) or higher ($q$-Laplacian).
Interestingly, binary classification problems are often approached through the learning of a continuous surrogate function $f$ whose sign is taken as the classification rule \cite{bartlett_convexity_2006}.
From a classification perspective, variations of $f$ are only needed in order for $f$ to change sign, and the low-variation hypothesis states that those variations should take place in sparsely populated areas of the input space.
This is coherent with the low-density separation hypothesis stating that $f$ should change sign in sparsely populated regions.

\subsection{Embedding for minimal variations}

This section extends on classical unsupervised techniques as aiming to minimize the criterion \eqref{eq:crit} under the orthonormal constraint \eqref{eq:const}.

Assuming low-variation of downstream tasks, it is natural to design $\phi$ in order to represent a maximum number of low-variation functions as $g\circ\phi$. 
Considering linear probes, the span of $(\phi_i)_{i\in[p]}$ could be searched as a $p$-dimension space of functions with minimal variations according to the criterion~\eqref{eq:crit}.
Put in equations with $\phi:\X\to\R^p$, this reads
\[
    \argmin_{\phi; \text{s.t. \eqref{eq:const}}}\max_{w\in\R^p; \norm{w} = 1} {\cal J}(w^\top \phi(\cdot)),
\]
under the coverage constraint \eqref{eq:const}.
Such a $\phi$ is an ideal data representation to solve downstream tasks with linear probes, as long as solutions verify the low-variation hypothesis.
With $D$ denoting the Jacobian, the formulation with $q=2$ translates as 
\begin{equation}
    \label{eq:lap}
    \phi = \argmin_{\phi\text{ s.t. \eqref{eq:const}}}\E\bracket{\norm{D \phi(X)}_F^2}.
\end{equation}
In practice, this formulation is favored for analytical reasons.
By making ${\cal J}(f)$ a quadratic form, it reveals the operator ${\cal L}$ that represents it.\footnote{Some minor mathematical precautions should be taken to deal with this weighted Sobolev pseudo-norm, we will omit them in this paper.}
In particular, equation \eqref{eq:lap} is solved explicitly with $\phi_i$ the $i$-th eigenfunctions of ${\cal L}$.\footnote{The solution of $\phi$ is unique up to orthonormal transformations $U\phi$ for $U\in\R^{p\times p}$ orthogonal, and to permutation of eigenfunctions associated with the $p$-th eigenvalue of ${\cal L}$.}
The proof is a simple application of the Rayleigh-Ritz formula, that defines eigenfunctions recursively through the formula
\begin{align*}
    &\phi_i = \argmin_{\phi:\X\to\R} \scap{\phi}{{\cal L}\phi} = \E\bracket{\norm{\nabla \phi(X)}^2}\\
    &\text{s.t. } \E\bracket{\phi_i(X) \phi_j(X)} = \delta_{ij} \quad \forall\, j < i.
\end{align*}

In the literature, this approach is often referred to as spectral embedding (the space is embedded through the spectral decomposition of the operator).
It is particularly well suited for the cluster assumption, since the null space of ${\cal L}$ is nothing but the span of the indicator functions of each connected component of $\rho_X$, which has motivated its use for clustering and manifold regularization \cite{belkin_manifold_2006}.
Under mild assumptions, ${\cal L}$ is indeed a diffusion operator (when $\rho_X$ has a density and compact support ${\cal L}f = -\Delta f + \scap{\nabla \log \rho_X}{\nabla f}$), which links it to diffusion maps \cite{coifman_diffusion_2006}, label propagation \cite{zhu_semi-supervised_2003} and Langevin dynamics \cite{bakry2014analysis}.

Since the 2000s, the criterion \eqref{eq:crit} has been approached in a non-parametric fashion based on finite differences, leveraging graph Laplacians \cite{zhu_semi-supervised_2003,belkin_laplacian_2003}.
Based on samples $(X_i)_{i\in[n]}$, it aims at minimizing
\begin{equation}
    \label{eq:finite}
    {\cal E}_g(\phi) = \sum_{i,j\in[n]}\bracket{k_\sigma(X_i, X_j) \norm{\phi(X_i) - \phi(X_j)}^2},
\end{equation}
with $k$ a notion of similarity to perform finite differences and $\sigma$ a scaling parameter ({\em e.g.} $k_\sigma(x, x') = \exp(-\norm{x-x'}^2/\sigma^2)$), and subject to the empirical version of the constraint \eqref{eq:const}, 
\[
    \frac{1}{n^2}\sum_{i, j \in [n]} \phi(X_i)\phi(X_i)^\top = I.
\]

\subsection{Consistency results}

This section discusses limiting behaviors of the methods described previously, namely SSL~\eqref{eq:ssl}, graph Laplacian~\eqref{eq:finite} and Dirichlet energy~\eqref{eq:lap}.

Graph Laplacians have arguably two convergence properties.
On the one hand, keeping the scale $\sigma$ constant, as the number of samples $n$ goes to infinity, the empirical minimizer minimizes the following measure of variations
\begin{gather*}
    \E\bracket{\norm{d(X)f(X) - k_\sigma*f(X)}^2}, \text{ with } 
    \\ k_\sigma * f(x) = \E[k_\sigma(x, X)f(X)],\quad d = k_\sigma * 1,
\end{gather*}
which can be seen as a smoothed, reweighted version of the Dirichlet energy.
The convergence happens relatively fast, typically in $O(n^{-1/2})$ in $L^2$-norm \cite{von_luxburg_consistency_2008}.
On the other hand, with the right scaling of $\sigma$, this finite difference method is able to converge towards the ideal solution defined by \eqref{eq:lap}.
Yet the convergence rates are much worse, {\em e.g.} in $O(n^{-1/d})$ for $d$ the dimension of the data manifold ($d=\dim\supp \rho_X$) \cite{hein_graph_2007}.
This may be understood intuitively, to measure variations with finite differences, the number of points needed grows exponentially with dimension \cite{bengio_label_2006}.

Alternatively, \eqref{eq:crit} might be estimated directly with empirical samples and a parametric model such as neural networks or kernel methods.
This enables fast convergence towards the solution of \eqref{eq:lap} within the search space of functions for $\phi$. 
By not suffering from the curse of dimension and converging to the ideal operator, this approach is statistically superior \cite{cabannes_overcoming_2021}.
Yet, it requires optimization over derivatives which can lead to computational drawbacks.

\begin{figure*}[t]
    \centering
    \includegraphics{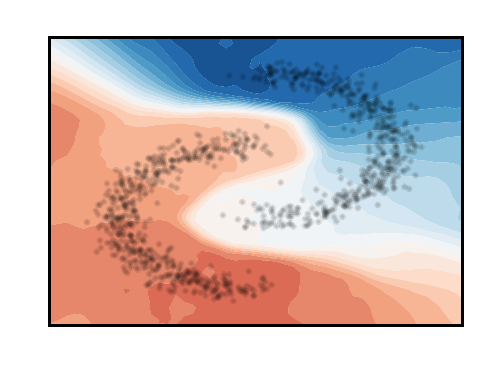}
    \includegraphics{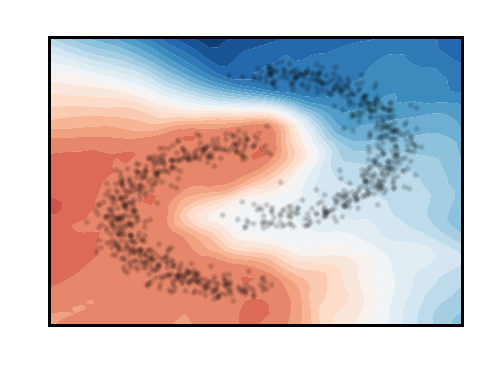}
    \includegraphics{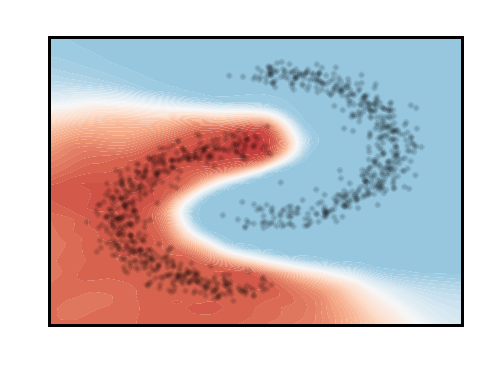}
    \includegraphics{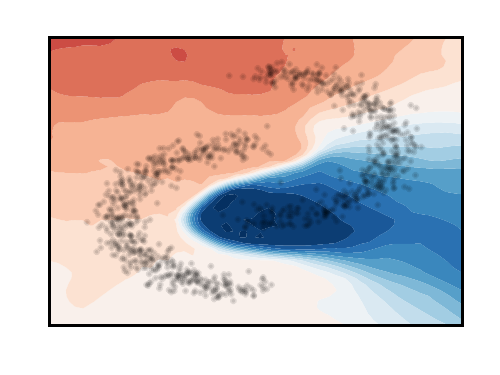}
    \includegraphics{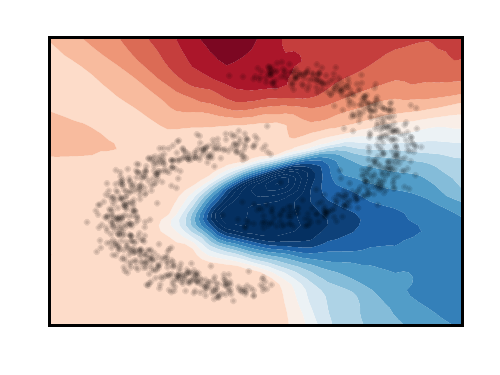}
    \includegraphics{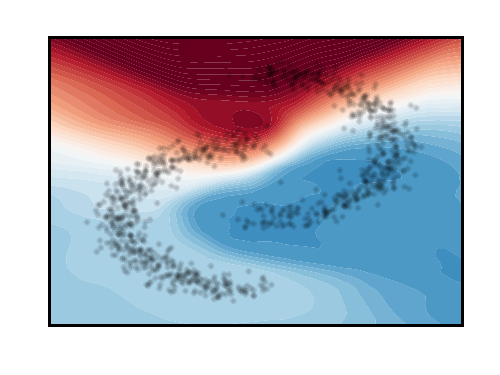}
    \caption{Functions learned with $p=2$ for SSL \eqref{eq:ssl} on the right, graph Laplacian \eqref{eq:finite} in the middle, and Dirichlet energy \eqref{eq:lap} on the right.
    Datapoints $(X_i)_{i\in[n]}$ are represented as black dots, while the functions $\phi_i(X)$ are represented through the level regions in different colors.}
    \label{fig:comp}
    \vspace{-.5em}
\end{figure*}

\subsection{Insights for SSL}
We argue that SSL objectives such as \eqref{eq:ssl} can be seen as measures of variations.
In particular, since $t(x, \xi)$ is supposed to be closed to $x$ (at least semantically speaking), it behaves as a random variable (with respect to $\xi$) to compute finite differences at a point $x\in\X$.
Therefore, we expect SSL algorithms either, when keeping the scale of $\xi$ constant,\footnote{Here, $\Xi$ is implicitly assumed to be a Banach space and the transformation to verify $t(x+\xi) = x + o(\norm{\xi})$.} to converge fast to the minimizer of some smoothed version of a functional that measure variations \eqref{eq:crit} (depending on $t$ and the distance $d$), or, with the right decreasing scale, to converge slowly to the ideal functional itself.

\section{Experiments}
\label{sec:exp}

Prior sections have introduced three techniques to learn $\phi$, SSL \eqref{eq:ssl}, graph Laplacian \eqref{eq:finite} and empirical Dirichlet energy \eqref{eq:lap}.
We have argued that they all aimed at learning the same type of functions, {\em i.e.} orthogonal functions that minimize variations.
After implementation details, proof-of-concept experiments verify this claim.

\subsection{Implementation details}
This section reviews implementation details based on empirical samples $(X_i)_{i\in[n]}$.
Experiments were made with the following specification of the self-supervised learning objective
\begin{equation}
    \tag{\ref{eq:ssl}}
    {\cal E}_s(\phi) = \frac{1}{n} \sum_{i\in[n]} \norm{\phi(X_i) - \phi(X_i + \sigma\xi_i)}^2,
\end{equation}
with $\xi_i$ a random unit Gaussian variable, and $\sigma$ a scale parameter.
The empirical version of Dirichlet energy reads
\begin{equation}
    \tag{\ref{eq:lap}}
    {\cal E}_e(\phi) = \frac{1}{n} \sum_{i\in[n]} \norm{\nabla\phi(X_i)}^2,
\end{equation}
which, in the case of deep networks, is related to double backpropagation~\cite{drucker1992improving} and has been used in other contexts~\cite{gulrajani_improved_2017,bietti2019kernel}.
Finally, the graph Laplacian objective is nothing but~\eqref{eq:finite}.

In experiments, the orthogonality constraints was relaxed as a penalty reading
\[
    \Omega(\phi) = \big\|\frac{1}{n^2}\sum_{i\in[n]} \phi(X_i)\phi(X_i)^\top - I\big\|^2,
\]
while the final objective is
\(
    {\cal E}(\phi) + \lambda \Omega(\phi).
\)

Self-supervised learning is known to be quite unstable to changes in hyperparameters.
In experiments, the scale parameters (standard deviation of augmentation in SSL, and kernel scaling in graph Laplacian) were set to match the width of the half-moon dataset (which was itself generated with Gaussian noise).
Stochastic gradient descent parameters (learning rate scheduling, batch size) were tuned to succeed the sole minimization of $\Omega$.\footnote{Note that because the expectation is inside the norm in $\Omega$ a naive mini-batch strategy does not provide unbiased stochastic estimate of its gradient. We overcame this issue by considering large batches.}
Finally, the regularizer $\lambda$ was set to approximately balance the penalty and the objective at hand (the learning rate was divided by $\lambda$ accordingly).
The representation $\phi$ was parameterized with a fully connected neural network with five hidden layers, each containing a hundred neurons.
The code is available online at \url{https://github.com/VivienCabannes/laplacian}.

\subsection{Consistency results.}
This section checks the claim that the three objectives \eqref{eq:ssl}, \eqref{eq:lap} and \eqref{eq:lap} are learning similar functions.
It proceeds with the two half-moons dataset (Figure~\ref{fig:comp}).

In this setting, the eigenfunctions of ${\cal L}$ are related to the Fourier basis on the union of two segments and are relatively stable under smoothing of the differential functional.
Beside the constant function, the null space of ${\cal L}$ is made of $\phi_1$ the difference of the indicator functions of both half-moons.
The second two eigenfunctions $\phi_2$ and $\phi_3$ are first-mode waves on each component.
Figure \ref{fig:comp} reports the learned $\phi$ for the three methods with $p=2$.
All methods recover $\phi_1$ (top), the first two recover $\phi_2$ while the third one recovers a mixture $\cos(\theta)\phi_2 + \sin(\theta)\phi_3$ for some $\theta\in[0, 2\pi]$, which also minimizes \eqref{eq:lap}.
Table \ref{tab:comp} is concerned with $p=5$, and the downstream task that consists in predicting if $x$ was in the top (or bottom) of the left (or right) half-moon.
This generates a classification problem with four different classes.
Such a task is ideal to evaluate our argumentation since the first five eigenfunctions of the diffusion operator ${\cal L}$ discriminate those four parts of the space with linear probing.
The results are satisfying.

\begin{table}[h]
    \centering
    \begin{tabular}{|c|c|c|}
    \hline
     SSL \eqref{eq:ssl} & energy \eqref{eq:lap} & graph \eqref{eq:finite}\\
     $96.14 \pm 0.16$ &  ${\bf 97.32} \pm 0.16$ & $95.15 \pm 0.19$\\
    \hline
    \end{tabular}
    \caption{Accuracy on downstream task with linear probing to check eigenspace retrieval (random is $0.25$).}
    \label{tab:comp}
    \vspace{-1em}
\end{table}

\subsection{Discussion}

While the previous experiments are made on small synthetic data, some behaviors are worse mentioning.
First, the SSL objective \eqref{eq:ssl} and the graph Laplacian \eqref{eq:finite} lead to similar results.
Yet, graph Laplacian only uses samples on the data manifold, while augmented data in SSL gets out of it.
At first sight, it seems better to restrict computations of finite differences to the manifold: the method would scale with the intrinsic dimension of data instead of the explicit input dimension \cite{hein_graph_2007}.
In practice, on the contrary, people do use aggressive color jittering leading to unnatural augmented images.
We notice in experiments that this prevents neural networks from taking arbitrary values outside the support of the data. 
On the other hand, graph Laplacian can exhibit high values outside the manifold, making it vulnerable to distribution shift or adversarial attacks \cite{hoffman_robust_2019}.\footnote{Additionally, rote that SSL gets fresh samples for each $\xi_i$ at each optimization epoch, which reduces in-samples bias.}

In high dimensional input space, the Dirichlet energy method \eqref{eq:lap} is supposed to exhibit much better statistical properties \cite{cabannes_overcoming_2021}. 
In practice, however, it suffers from some computational drawbacks.
More specifically, for neural networks with two hidden layers with both one hundred neurons, graph Laplacian and SSL find similar solutions as the one in Figure \ref{fig:comp} while the Dirichlet energy method tends to collapse to basic orthogonal functions such as $\phi_i = \cos(2\pi \omega_i\scap{e_i}{x})$ for some small $\omega$ and some unit vector $e_i$.
This behavior vanishes with deeper networks.

Finally, when $p$ gets big, the different functions $\phi_i$ learned are hard to parse visually. 
While a solution for $\phi$ are waves with increasing modes, in practice the networks learn an orthogonal transformation of it, {\em i.e.} $\phi \leftarrow U\phi$ for $U\in\R^{p\times p}$ a random orthogonal matrix.
If those different modes were to correspond to features in the original data, it would be natural to ask for $(\phi_i)$ to describe those local regions of the input space associated with features.
This suggests room for future improvements of SSL methods.

\section{Conclusion}
This paper unveiled the link between novel self-supervised learning techniques and classical unsupervised learning ones.
Key to all those methods is the low-variation hypothesis.
In future work, we hope to leverage this understanding to provide practical guidelines to design self-supervised learning algorithms and deploy them in the wild without having to rely on expensive hyperparameters validation.

\bibliographystyle{template/IEEE}
\bibliography{main}

\end{document}